\documentclass[lettersize,journal]{IEEEtran}
\usepackage{amsmath,amsfonts}
\usepackage{algorithmic}
\usepackage{algorithm}
\usepackage{array}
\usepackage{xcolor}
\usepackage[caption=false,font=normalsize,labelfont=sf,textfont=sf]{subfig}
\usepackage{textcomp}
\usepackage{pifont}
\usepackage{stfloats}
\usepackage{url}
\usepackage{verbatim}
\usepackage{graphicx}
\usepackage{cite}
\usepackage{multirow}
\begin{document}

\title{Mix-QViT: Mixed-Precision Vision Transformer Quantization Driven by Layer Importance and Quantization Sensitivity }

\author{Navin Ranjan and Andreas Savakis \\
Rochester Institute of Technology\\ Rochester, New York 14623, USA

}



\maketitle

\begin{abstract}
Vision transformers (ViTs) have achieved remarkable performance across various visual tasks. However, these models face significant computational and memory demands, making them difficult to deploy on resource-constrained platforms. Quantization is a widely adopted technique for reducing model size, but most existing studies focus primarily on uniform bit-width quantization across the entire network, which often leads to suboptimal performance. Although some studies have explored mixed-precision quantization (MPQ) for ViTs, these methods typically rely on exhaustive search algorithms or arbitrarily assign different bit-widths without systematically considering each layer’s specific role and sensitivity to quantization. In this paper, we propose Mix-QViT, an explainability-driven MPQ framework that systematically allocates bit-widths to each layer based on two criteria: layer importance, assessed via Layer-wise Relevance Propagation (LRP), which identifies how much each layer contributes to the final classification, and quantization sensitivity, determined by evaluating the performance impact of quantizing each layer at various precision levels while keeping others layers at a baseline. Additionally, for post-training quantization (PTQ), we introduce a clipped channel-wise quantization method designed to reduce the effects of extreme outliers in post-LayerNorm activations by removing severe inter-channel variations. We validate our approach by applying Mix-QViT to ViT, DeiT, and Swin Transformer models across multiple datasets. Our experimental results for PTQ demonstrate that both fixed-bit and mixed-bit methods outperform existing techniques, particularly at 3-bit, 4-bit, and 6-bit precision. Furthermore, in quantization-aware training, Mix-QViT achieves superior performance with 2-bit mixed-precision.
\end{abstract}


\section{Introduction}
\label{sec:intro}
\IEEEPARstart{V}{ision} Transformers (ViTs) have demonstrated state-of-the-art (SOTA) performance across various vision tasks, including image classification~\cite{image_16x16_dosovitskiy2021an_ViT, liu2021swin, wang2021PVT}, object detection~\cite{mao2022exploring_OD, carion2020end_end_OD} and segmentation~\cite{zheng2021rethinking_seg,strudel2021segmenter_seg}. However, their high computational demands, substantial memory footprint, and significant energy consumption make them impractical for deployment on resource constrained platforms. 
To tackle these challenges, numerous compression and acceleration techniques have been proposed, such as  pruning~\cite{he2018soft_prun, yu2022width_prun}, low-rank decomposition~\cite{denil2013predicting_lw}, quantization~\cite{lin2022fq_vit, li2023repq, li2022QViT}, knowledge distillation~\cite{chen2022dear_KD, lin2022knowledge_KD}, dynamic token reduction~\cite{rao2021dynamicvit, xu2022evo_token}, and more compact ViT architecture~\cite{rao2021dynamicvit, li2022efficientformer, wu2022tinyvit, zhang2022minivit}. All of these methods aim to reduce model size and complexity while preserving performance. 

Model quantization is particularly popular for reducing both model size requirements and computational costs by representing network weights and activations with low-bit precision. However, quantization often leads to significant performance drops. To mitigate these losses, researchers have explored two main quantization paradigms: Quantization-Aware Training (QAT) and Post-Training Quantization (PTQ). QAT retrains the model with low-bit precision on the entire dataset to recover the lost performance but requires significant training time and computational resources. 
In contrast, PTQ directly quantizes a pre-trained model using a small calibration set, making it faster and more practical, though it generally incurs larger performance degradation compared to QAT.

Recent PTQ studies on ViTs~\cite{ding2022APQViTtowards, yuan2022ptq4vit, lin2022fq_vit, li2023repq, AdaLog_wu2025} have identified LayerNorm, Softmax, and GELU as highly sensitive to quantization, often resulting in significant performance degradation. To address these bottlenecks, several targeted enhancements have been proposed. 
For example, APQ-ViT~\cite{ding2022APQViTtowards} uses block-wise calibration with Matthew-effect-preserving quantization, while FQ-ViT~\cite{lin2022fq_vit} introduces the Power-of-Two Factor to handle inter-channel variation in LayerNorm and employs Log-Int-Softmax for Softmax layer quantization. RepQ-ViT~\cite{li2023repq} and AdaLog~\cite{AdaLog_wu2025} implement a decoupled quantization-inference framework. During calibration, both frameworks utilize channel-wise quantization for LayerNorm. For the Softmax layer, RepQ-ViT applies logarithmic quantization with a base of $\sqrt{2}$, while AdaLog uses adaptive-base logarithmic quantization to enhance representational capacity. In the inference stage, both frameworks transition to hardware-efficient formats, adopting layer-wise quantization for LayerNorm and log2 quantization for the Softmax layer. Despite these advancements, significant research gaps remain. Current methods still show a large performance gap between low-bit (e.g., 4-bit) models and full-precision counterparts, with the gap becoming more pronounced 
under ultra-low-bit configurations, e.g. 3 bits. 

Moreover, most existing QAT and PTQ methods apply uniform bit precision across the entire network, assuming equal importance and quantization sensitivity for all the layers. This uniform approach is often suboptimal, as it forces critical layers to operate at the same precision as less significant ones, thereby missing opportunities to reduce model size and improve performance. Mixed-precision quantization (MPQ) addresses this limitation by allowing different layers to operate at different bit widths, preserving performance through higher precision in crucial layers while maintaining efficiency with fewer bits elsewhere. Prior MPQ studies employ either search-based methods~\cite{liu2021post_Rank_Aware_PTQ, lou2020autoq, xiao2023patch-wise} or criterion-based methods~\cite{dong2019hawq, dong2020hawqv2} to optimize bit precision for individual layers, however, they are often computationally expensive or are lacking interpretability. For instance, the work in~\cite{liu2021post_Rank_Aware_PTQ} uses nuclear norm metrics for Multi-Head Self-Attention (MHSA) and MLP modules to measure sensitivity, while PMQ~\cite{xiao2023patch-wise} measures a layer's sensitivity by evaluating the error induced by removing each layer. Hessian-based methods~\cite{wang2019haq, dong2019hawq, dong2020hawqv2, yang2023global_nvit, CLADO_deng2023mixed} provide another avenue to measure importance score but frequently incur high computational overhead and fail to explain why certain layers deserve more precision.

In this paper, we address these gaps by proposing a \textbf{mixed-precision quantization strategy for ViTs} guided by two complementary scores: 
{\em 
(i) an  explainability-based layer importance score, and (ii) a quantization sensitivity score. 
}
The layer importance score, derived using Layer-wise Relevance Propagation (LRP)~\cite{ chefer2021LRP}, provides an interpretable measure of how much each layer contributes to the final prediction. On the other hand, the quantization sensitivity score captures how each layer’s performance changes under different bit configurations relative to a baseline. 
By integrating these two metrics, we formulate an Integer Quadratic Problem (IQP) that allocates optimal bit-widths to each layer under constraints, such as model size and bit operations. This unified framework enables an explainable allocation of bits, addressing the limitations of search-based approaches that lack interpretability and Hessian-based methods that require substantial computation.

Additionally, we enhance existing PTQ frameworks, such as RepQ-ViT~\cite{li2023repq} and AdaLog~\cite{AdaLog_wu2025}, by introducing clipped channel-wise quantization for post-LayerNorm activations during both calibration and inference, to handle the inter-channel variation. This modification mitigates performance degradation seen at ultra-low bit widths (3-4 bits) when reparameterizing from channel-wise to layer-wise scales. We also adopt the Softmax quantization improvements from~\cite{li2023repq, AdaLog_wu2025} to further preserve accuracy in ViT models. Our main contributions are as follows:
\begin{itemize}
    \item We propose Mix-QViT, a mixed-precision framework for quantizing vision transformers.
    \item Our mixed-precision quantization strategy relies on two key criteria: \textbf{layer importance score}, which allocates higher precision to critical layers and lower bits to less significant ones, and \textbf{quantization sensitivity score}, which adjusts precision to reduce the impact on sensitive layers. These objectives are formulated as an Integer Quadratic Problem (IQP) to optimize mixed-bit allocations under model size and computational constraints.
    \item Mix-QViT introduces a novel method for quantifying layer importance using Layer-wise Relevance Propagation (LRP). Unlike~\cite{chefer2021LRP}, which relies on CLS tokens, we use image tokens to generate relevance maps, making the method applicable to all transformer architectures. A layer's contribution score is calculated as the average of mean relevance values across inputs, and the layer importance score is derived by normalizing each contribution score relative to the total contributions across all layers.
    \item We introduce Quantization Sensitivity Analysis (QSA) to assess each layer's robustness to quantization. Starting with a baseline configuration where all layers use the same precision, we systematically vary the precision of each layer within a predefined range while keeping others fixed at the baseline. The model's performance is measured after each adjustment, and the quantization sensitivity score is calculated as the average normalized performance change across layers and precision levels.
    \item For PTQ, we introduced a clipped channel-wise quantization method for post-LayerNorm activations during both calibration and inference. This method effectively mitigates inter-channel outliers, improving performance for both fixed-bit and mixed-precision quantization, particularly in ultra-low bit scenarios (3-4 bits).
    \item Extensive experiments on ViT, DeiT, and Swin models demonstrate that Mix-QViT consistently outperforms existing methods. It achieves superior results across classification, detection, and segmentation benchmarks for PTQ at 3-, 4-, and 6-bit fixed and mixed-bit precision, as well as in classification benchmarks for QAT with 2-bit mixed precision.
\end{itemize}

The paper is structured as follows: Section~\ref{sec:methodology} reviews the background and introduces the proposed solutions, including clipped channel-wise quantization, layer sensitivity scores, quantization sensitivity scores, and mixed-precision bit allocation. Section~\ref{sec:result} presents results and key findings across multiple benchmarks for both PTQ and QAT. Finally, Section~\ref{sec:conclusion} summarizes the contributions and key insights. 

\section{Methods}
\label{sec:methodology}

\subsection{Vision Transformer Architecture}
The vision transformer reshapes an input image into a sequence of \(N\) flattened 2D patches, each mapped to a \(D\)-dimensional vector through a linear projection layer, denoted by $X\in{R}^{N\times D}$. The ViT consists of multiple transformer blocks, each with a multi-head self-attention (MHSA) module and a multi-layer perceptron (MLP) module. The MHSA module captures token relationships for global feature extraction. In the \(l^{th}\) block, the input \(X^l\) is projected to generate query, key, and value matrices: \(Q_l = X^lW^q_i\), \(K_l = X^lW^k_i\), and \(V_l = X^lW^v_i\), abbreviated as QKV. Attention scores are computed between the queries and keys, followed by a softmax operation, which is used to weight the values. The MHSA output is then generated by concatenating the weighted values from all attention heads, as follows:
\begin{align}
\text{Attn}_{i} = \text{Softmax}\left(\frac{Q_{i}\cdot K_{i}^T}{\sqrt{D_{h}}}\right)V_{i},
\label{eq:attn} \\
\text{MHSA}(X^{l}) = [\ \text{Attn}_{1}, \text{Attn}_{2},...,\text{Attn}_{i}]\ W^o,
\label{eq:attn_out}
\end{align}
where \(h\) is the number of attention heads, \(D_{h}\) is the feature size of each head, and \(i = 1, 2, \dots, h\). The MLP module consists of two fully connected layers separated by a GELU activation, projecting features into a high-dimensional space, as follows:
\begin{equation}
\label{eq:mlp}
\text{MLP}(Y^{l}) = \text{GELU}\left(Y^{l}W^{1} + b^{1} \right)W^{2} + b^{2}.
\end{equation}
Here, \(Y^{l}\) is the MLP input, \(W^{1} \in \mathbb{R}^{D \times D_f}\), \(b^{1} \in \mathbb{R}^{D_f}\), \(W^{2} \in \mathbb{R}^{D_f \times D}\), and \(b^{2} \in \mathbb{R}^{D}\). Layer normalization (\(LN\)) precedes each module, with residual connections added after. The transformer block is defined as:
\begin{align}
Y^{l} = X^{l} + \text{MHSA}\left(LN\left(X^{l}\right)\right),
\label{eq:mhsa_module} \\
X^{l+1} = Y^{l} + \text{MLP}\left(LN\left(Y^{l}\right)\right).
\label{eq:mlp_module}
\end{align}
The large matrix multiplications in MHSA and MLP significantly contribute to computational costs. To address this, we apply quantization to weights and activations across layers, including linear layers (QKV, Projection, FC1, FC2, and classifier head), the convolutional layer (Patch Embedding), and matrix multiplications (MatMul1, MatMul2), using a uniform quantizer. For power-law-like distributions, such as those in Post-Softmax and Post-GELU layers, we utilize logarithmic quantization with fixed bases (\(\log_2\), \(\log\sqrt{2}\)) or an adaptive base as proposed in~\cite{AdaLog_wu2025}.

\subsection{Model Quantization}
The uniform quantization divides the data range into equal intervals and is defined as: 
\begin{align}
\text{Quant: } x^{q} &= \text{clip}\left(\left\lfloor\frac{x}{s}\right\rceil + z, 0, 2^{b}-1\right) \label{eq:uniform_quant} \\
\text{DeQuant: } \bar{x} &= s \cdot (x^{q} - z) \approx x \label{eq:uniform_qequant} \\
s = \frac{\max(x) - \min(x)}{2^b-1}, &\quad \text{and} \quad 
z = \left\lfloor -\frac{\min(x)}{s} \right\rceil \label{eq:scale_zeropoint}
\end{align}
Here, \(x\) represents the original floating-point weights or inputs, \(x^{q}\) is the quantized value, \(b\) is the quantization bit precision, \(\lfloor\cdot\rceil\) denotes rounding, and \(\text{clip}\) ensures values remain within the quantization range. The quantization scale \(s\) and zero-point \(z\) are determined by the lower and upper bounds of $x$. During dequantization,  \(\bar{x}\) approximately recovers the original \(x\).

Logarithmic quantization, particularly \(\log_2\), is a widely used approach for handling power-law distributions. Recent studies have introduced a quantization-inference decoupling paradigm, where various logarithmic bases are explored during the quantization phase to enhance model performance~\cite{li2023repq, AdaLog_wu2025}. These bases are later reparameterized to \(\log_2\) to ensure efficient and hardware-friendly inference. The generalized logarithmic quantization is defined as:
\begin{align}
\text{Quant: } x^{q}&= \text{clip}\left( 
\left\lfloor -\log_{a} \frac{x}{s} \right\rceil, 0, 2^{b} - 1 \right) \notag\\
 &= \text{clip}\left(\left\lfloor -\frac{\log_2 \frac{x}{s}}{\log_2 a} \right\rceil, 0, 2^{b} - 1 \right) \label{eq:generalized_log_quant} \\
\text{DeQuant: } \bar{x} &= s \cdot a^{-x^{q}} \approx x \label{eq:generalized_log_dequant}
\end{align}
For a fixed base \(a = 2\), the method represents standard \(\log_2\) quantization. With a fixed base \(a = \sqrt{2}\), it corresponds to \(\log\sqrt{2}\) quantization, as proposed in~\cite{li2023repq}. For adaptive bases, as introduced in~\cite{AdaLog_wu2025}, \(a\) is dynamically optimized during training to enhance performance.

\subsection{Clipped Channel for LayerNorm Activations}
In ViTs, during inference, LayerNorm computes the statistics $\mu_x$, and $\sigma_x$ at each forward step and normalizes the input $X\in\mathbb{R}^{N\times D}$. The affine parameters $\gamma\in\mathbb{R}^D$ and $\beta\in\mathbb{R}^D$ then re-scale the normalized input to another learned distribution. The LayerNorm process is defined as follows:
\begin{equation}
\text{LayerNorm(X)}=\frac{X-\mu_x}{\sqrt{{\sigma_x^{2}}+\epsilon}}\odot \gamma + \beta,
\label{eq:LayerNorm_def}
\end{equation}
where $\odot$ denotes the Hadamard product.
\begin{figure}[t]
    \centering
    \subfloat[\label{fig:blocks.0.attn.qkv}]{
        \includegraphics[width=0.23\textwidth]{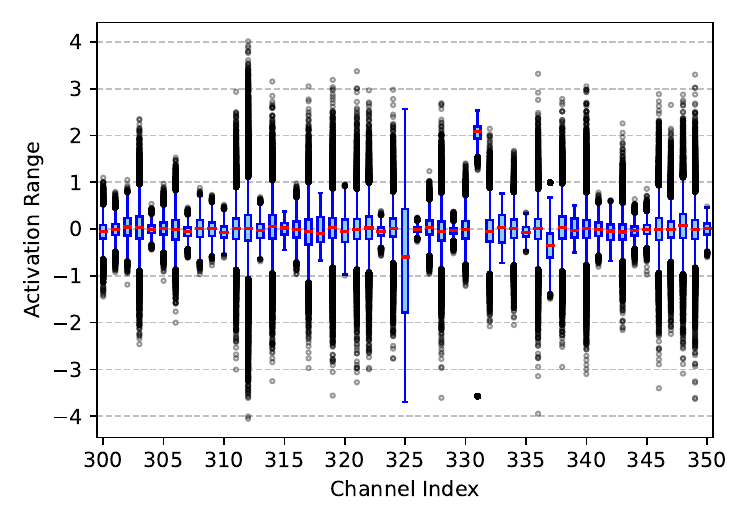}
    }%
    \hfill
    \subfloat[\label{fig:blocks.7.mlp.fc1}]{
        \includegraphics[width=0.23\textwidth]{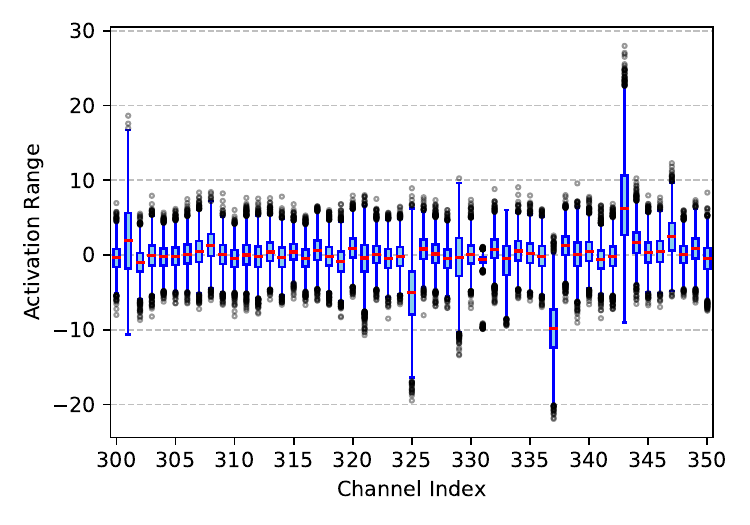}
    }%
    \caption{Box plot of the post-LayerNorm activations for channels 300 to 350 in DeiT-S: (a) activations from the blocks.0.attn.qkv layer, and   (b) activations from the blocks.7.mlp.fc1 layer. Extreme values are marked with black circle.}
    \label{fig:LayerNorm_inter-channel_variation}
\end{figure}

Analyzing Post-LayerNorm activations reveals substantial inter-channel variation, as shown in Fig.~\ref{fig:LayerNorm_inter-channel_variation}, which hinders the effectiveness of post-training quantization. To address this, several strategies have been proposed. Group-wise quantization~\cite{shen2020qQBert} assigns separate quantization parameters to matrices corresponding to individual attention heads in MHSA. Channel-wise quantization~\cite{lin2022fq_vit} applies distinct parameters to each channel using a power-of-two factor approach. In~\cite{li2023repq, AdaLog_wu2025}, a quantization-inference decoupling method was introduced, which performs channel-wise quantization during training and reparameterizes these parameters into layer-wise quantization by averaging during inference.


Building on the works of RepQ-ViT~\cite{li2023repq} and AdaLog~\cite{AdaLog_wu2025}, we introduce Clipped Channel-wise Reparameterization for LayerNorm activations (CRL) to enhance inference performance. By clipping outliers within a few standard deviations (\( \sigma\)) from the channel mean, CRL minimizes inter-channel variation and significantly improves performance compared to~\cite{li2023repq, AdaLog_wu2025}, particularly at ultra-low precisions such as 3 and 4 bits, though with a small throughput loss, as detailed in Table~\ref{tab:Efficiency_analysis_Data_Time_Throughput}. We integrate CRL into RepQ-ViT, referred to as RQViT, and into AdaLog, referred to as AQViT.

For the \(l^{th}\) transformer block, given the input \(X^{l}_{LN}\), we perform channel-wise quantization to derive the scale \(s \in \mathbb{R}^D\) and zero-point \(z \in \mathbb{R}^D\). We then compute the clipped channel-wise scale \(\hat{s}\) and zero-point \(\hat{z}\) as:
\begin{align}
\hat{s} &= \text{clip}(s, \mu_s - 2\sigma, \mu_s + 2\sigma), \notag \\
\hat{z} &= \text{clip}(z, \mu_z - 2\sigma, \mu_z + 2\sigma),
\label{eq:scale_and_zeropoint_clip} 
\end{align}
where \(\mu_s\) and \(\mu_z\) are the means, and \(\sigma\) is the standard deviation of \(s\) and \(z\), respectively. Through experimentation, we discovered that using clipping at 2\(\sigma\) achieves optimal performance. The variation factors between the original and clipped parameters are denoted as \(v_1 = s / \hat{s}\) and \(v_2 = z - \hat{z}\). Using these, (\ref{eq:scale_and_zeropoint_clip}) can be expressed as:
\begin{align}
\hat{s} &= \frac{s}{v_1} = \frac{\left[\max(x) - \min(x)\right]/v_1}{2^b - 1}, \label{eq:scale_variation} \\
\hat{z} &= z - v_2 = \left\lfloor -\frac{\min(x) + s \odot v_2}{s} \right\rceil. \label{eq:zeropoint_variation}
\end{align}
In (\ref{eq:scale_variation}), dividing each channel of $X^l_{LN}$ by $v_1$ results in $\hat{s}$. Similarly, in Eq. (\ref{eq:zeropoint_variation}), adding $s\odot v_2$ to each channel of $X^l_{LN}$ gives $\hat{z}$. These operations can be achieved by adjusting the LayerNorm's affine factors as follows:
\begin{align}
\hat{\beta} = \frac{\beta+s\odot v_2}{v_1},
\text{\hspace{0.1in}}
\hat{\gamma} =\frac{\gamma}{v_1}.
\label{eq:affine_factors}
\end{align}

This reparameterization induces a change in the activation distribution, specifically expressed as $\hat{X}^l_{LN} = \left(X^l+s\odot v_2\right)/v_1$. 
In the MHSA module, the layer following LayerNorm is a linear projector layer for QKV. The reparameterized shift in the QKV layer is expressed as
\begin{align}
X^l_{LN}\cdot W^{QKV} = \frac{X^l_{LN}+s\odot v_2}{v_1}\left(v_1\odot W^{QKV}\right) \notag \\+ \left(b^{QKV}-\left(s\odot v_2\right)W^{QKV}\right).
\label{eq:repram_QKV}
\end{align}
Here, $W^{QKV}\in \mathbb{R}^{D\times D_{h}}$ and $b^{QKV} \in \mathbb{R}^{3D_h}$ represent the weight and bias of the QKV layer. To offset this distribution shift, the weight of the subsequent layer can be adjusted, as outlined below:
\begin{align}
\hat{W}^{QKV} = v_1\odot W^{QKV}, \notag \text{\hspace{0.5in}}
\\ 
\hat{b}^{QKV} = b^{QKV}-\left(s\odot v_2\right)W^{QKV}.
\label{eq:weight_offset_QKV}
\end{align}
This strategy is similarly applied to the inputs of the FC1 layer in the MLP module. 

\subsubsection{Quantization for power-law-like distribution Activations}
In ViTs, the softmax operation in the MHSA module produces a highly imbalanced power-law distribution, with most values being small and a few having large magnitudes, making it unsuitable for quantization. Existing methods~\cite{lin2022fq_vit, ding2022APQViTtowards} have improved upon uniform quantization but often fail to consistently deliver satisfactory results. In~\cite{li2023repq}, a quantization-inference decoupling approach employs a \(\log\sqrt{2}\) quantizer for softmax activation, offering higher resolution and better representation of the power-law distribution. During inference, this quantizer is reparameterized into a hardware-friendly \(\log_2\) quantizer, balancing accuracy and efficiency. Similarly,~\cite{AdaLog_wu2025} proposes a progressive search strategy to dynamically optimize the logarithmic base for post-Softmax and post-GELU activations, enabling hardware-friendly quantization and dequantization. In this work, both of our PTQ frameworks (RQViT and AQViT) incorporate their respective counterpart's modifications to effectively handle post-Softmax and post-GELU activations.

\begin{figure*}[t]
\centering
  \includegraphics[width=1.0\linewidth]{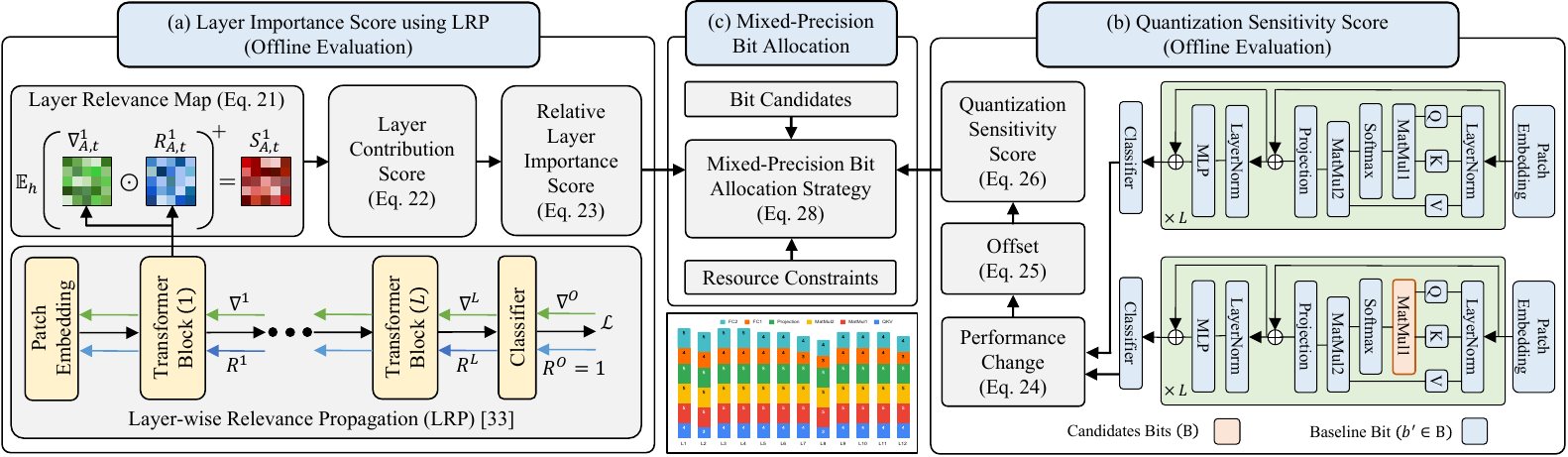}
  \caption{Overview of the Mix-QViT framework.
  (a) Layer Importance Score (\(\Omega\)), calculated offline on a small sample of 256 images from the ImageNet1K validation dataset, based on the Layer-wise Relevance Propagation (LRP) method. (b) Quantization sensitivity score (\(\Lambda\)), calculated offline on 256 images from the ImageNet1K validation dataset. Here, performance changes are recorded between two models: one where all layers are quantized at baseline precision, and another where the target layer in each transformer block is quantized at different precision. (c) Mixed-precision bit allocation strategy, based on the layer importance score and quantization sensitivity score, which, under model constraints, generates an optimal mixed-bit allocation. 
  }
  \label{fig:Mix-QViT Framework}
\end{figure*}

\subsection{Layer Importance Score using LRP}
Layer-wise Relevance Propagation (LRP) is an explainability technique that uses gradients and relevance scores to uncover how individual features contribute to a model's output. For a classification task with \( C \) classes and an input image corresponding to class \( c \), where \( c \in [1, \dots, C] \). Let \( x^{(l)} \) denote the input to the \( l \)-th layer of a network with \( L \) layers, where \( x^{1} \) represents the network input and \( x^{L} \) corresponds to its output. The gradient of the \( l \)-th layer with respect to the classifier's output \( y_c \) is given by:  
\begin{equation}
\label{eq:gradient_flow}
\nabla x^{l}_{j} = \frac{\partial y_c}{\partial x^{l}_{j}} = \sum_{i} \frac{\partial y_c}{\partial x^{l+1}_{i}} \cdot \frac{\partial x^{l+1}_{i}}{\partial x^{l}_{j}},
\end{equation}  
where the index \( j \) corresponds to elements in \( x^l \), and \( i \) corresponds to elements in \( x^{l+1} \).  

Similarly, relevance \( R^o = 1_c \), where \( 1_c \) is a one-hot vector indicating the target class \( c \), is propagated backward from the output layer through the network to the input. Relevance propagation in LRP is based on Deep Taylor Decomposition~\cite{montavon2017explainingDTD}. Let \( M^l_i(X^l, W^l) \) represent the operation performed by layer \( l \) on the input feature map \( X^l \) with weights \( W^l \). The relevance at layer \( l \) is propagated as:
\begin{align}
R^{l}_{j} &= \mathcal{G}(X^l, W^l, R^{l+1}) \notag \\  
&= \sum_{i} x^l_j \frac{\partial M^l_i(X^l, W^l)}{\partial x^l_j} \frac{R^{l+1}_i}{M^l_i(X^l, W^l)}.
\label{eq:relevance}
\end{align}

The transformer block uses GELU activation~\cite{hendrycks2016GELU}, which produces both positive and negative values. To address potential inaccuracies in relevance scores from these mixed values, we follow~\cite{chefer2021LRP} and exclude negative elements by defining a subset of indices, \( p = \{(i,j) \mid x_jw_{ji} \geq 0\} \). The relevance propagation is then expressed as:
\begin{align}
R^{l}_{j} &= \mathcal{G}_{p}\left(x,w,p,R^{l+1}\right) \notag \\
&= \sum_{\{\left(i\mid i,j\right)\in p\}}\frac{x_jw_{ji}}{\sum_{\{j'|\left(j',i\right)\in p\}}x_{j'}w_{j'i}}R^{l+1}_i.
\label{eq:positive_relevance}
\end{align}

The LRP method~\cite{chefer2021LRP} computes relevance score maps for the attention layer in the MHSA module by performing a Hadamard product between the relevance of the CLS token and its gradient.
However, this approach is restricted to architectures that utilize a CLS token. In this work, we generalize relevance mapping to all layers of the transformer and extend it to any transformer architecture by utilizing image tokens instead of the CLS token. For \( t \)-th sample, the attention layer \( A \) in transformer block \( L \), with gradients \( \nabla A^L_t \) and relevance \( R^L_{A,t} \), the relevance score map \( S^L_{A,t} \) is defined as:
\begin{equation}
\label{eq:relevance_map}
{S}^{L}_{A,t} = \mathbb{E}_h\left(\nabla_{A,t}^L \odot R^L_{A,t}\right)^+.
\end{equation}
Here, $\mathbb{E}_h$ represents the mean across the attention heads, and our analysis focuses solely on the positive values  from the relevance map. 

To quantify the contribution of attention layer for a given sample image toward the model's output, we compute the mean of its relevance map. Since the LRP method is class-specific, each sample generates a unique relevance map. Consequently, the overall contribution score is obtained by averaging the relevance score across \(T = 256\) randomly selected images from the ImageNet1k validation dataset, as:
\begin{equation}
\label{eq:contribution_score}
{C}^{L}_A = \frac{1}{T}\sum_{t=1}^{T}\mathbb{E}\left({S}^{L}_{A,t}\right).
\end{equation}
Further details on the impact of sample size on the contribution score and its influence on model quantization performance are discussed in Table~\ref{tab:Data Size & Component Ablation Study}. This approach is similarly applied to all other quantized layers (\(U\)).
The relative layer importance score (\(\Omega\)) for any layer in any block is defined as the average normalized contribution score of that layer relative to all contribution scores. For the attention layer in the \(L\)-th block, the relative \textbf{layer importance score} is expressed as:
\begin{equation}
\label{eq:layer_importance_score}
{\Omega}^{L}_A = \frac{C^L_A}{\sum_{l=1}^{L}\sum_{u \in U_l}C^l_u}.
\end{equation}
Here, \(U_l\) denotes all the quantized layers in the \(l\)-th transformer block. 
\begin{figure}[t]
\centering
  \includegraphics[width=0.7\linewidth]{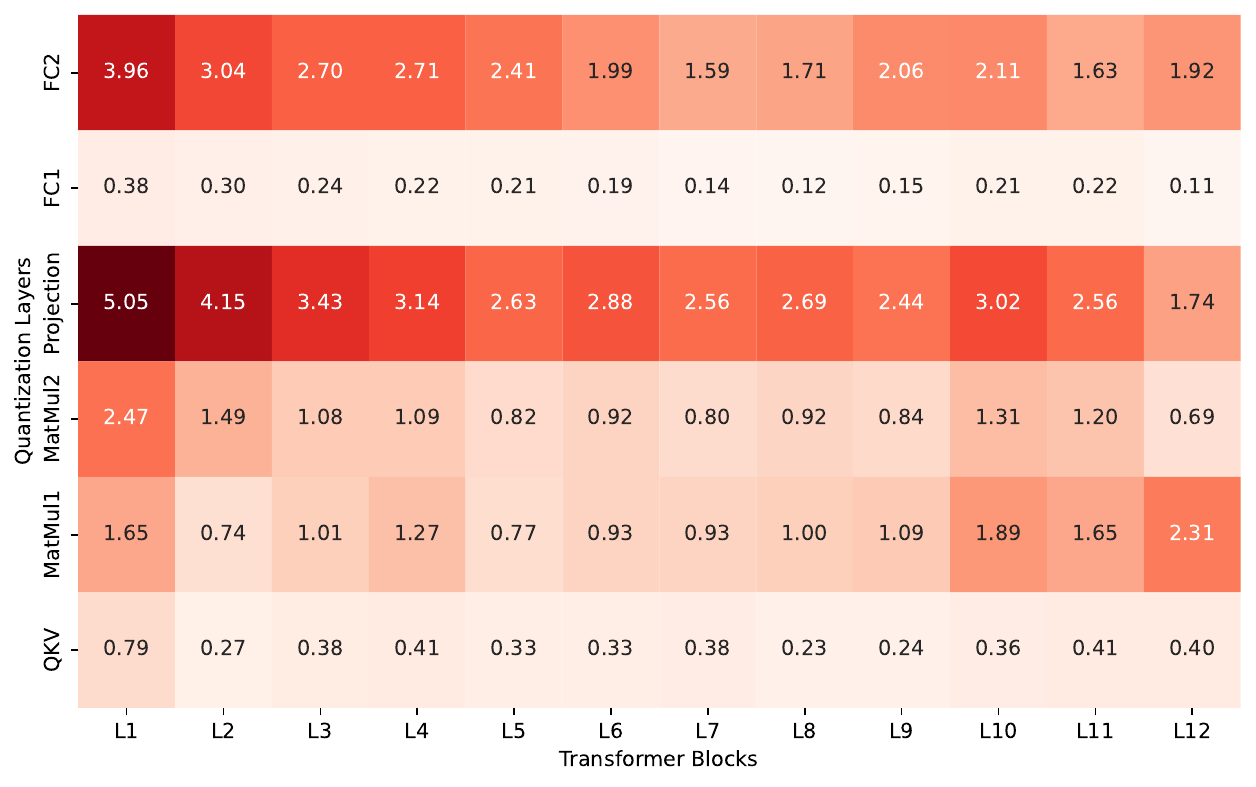}
  \caption{Layer Importance Score of DeiT-Small. Each value refers to the layer relative importance toward model classification. }
  \label{fig:Layer Importance score for DeiT-Small}
\end{figure}

Fig. \(\ref{fig:Layer Importance score for DeiT-Small}\) shows the importance scores for quantized layers across transformer blocks in the DeiT-Small model, highlighting their contribution to output classification. The Projection layer consistently has the highest scores, underscoring its critical role and the need for higher precision during quantization. In contrast, the FC2 layer exhibits lower scores, suggesting it can tolerate reduced bit-widths with minimal accuracy loss. The QKV and MatMul1 layers show moderate importance, with higher scores in earlier blocks and lower scores in later ones, indicating potential for further bit-width reductions in deeper layers. This analysis emphasizes the importance of quantization strategies guided by layer significance.

\subsection{Quantization Sensitivity Analysis}  
Prior research on transformer quantization sensitivity often suffers from limited scope or computational efficiency. For instance,~\cite{Q-ViT_2_Q_sensitivity} used a PTQ approach to analyze the MLP layers comprehensively but only evaluated MHSA at head levels, omitting individual layer analysis. Similarly, ~\cite{Quantization_variation} employed a leave-one-out method to assess query, key, and value matrices but overlooked other MHSA layers. Moreover, this study relied on computationally expensive QAT, limiting its practicality.

\begin{figure}[t]
    \centering
    \includegraphics[width=0.4\textwidth]{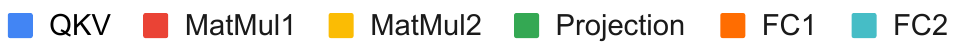}
    
    \hfill
    \subfloat[\label{fig:Quantization_Sensitivity_acc_drop_A}]{
        \includegraphics[width=0.23\textwidth]{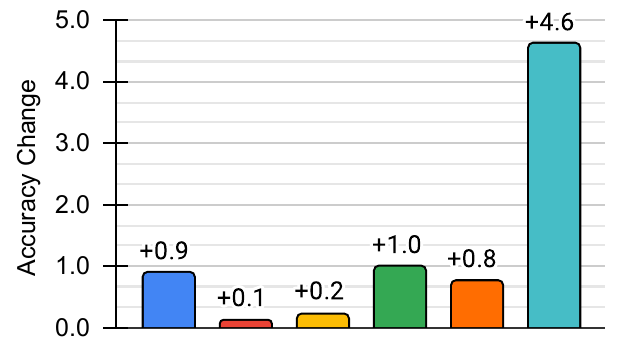}
    }
    \hfill
    \subfloat[\label{fig:Quantization_Sensitivity_acc_drop_B}]{
        \includegraphics[width=0.23\textwidth]{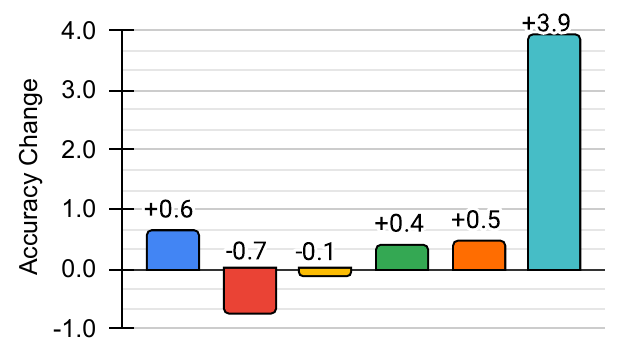}
    }
    \hfill
    \subfloat[\label{fig:Quantization_Sensitivity_acc_drop_C}]{
        \includegraphics[width=0.23\textwidth]{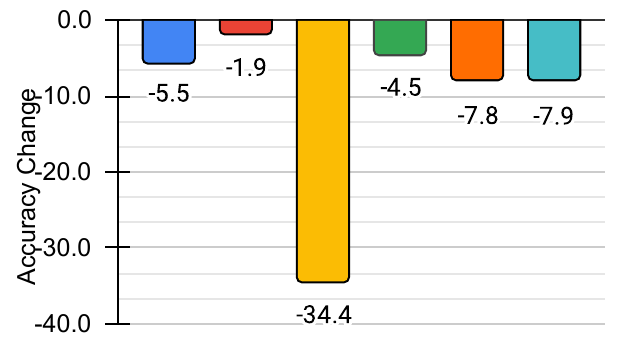}
    }
    \hfill
    \subfloat[\label{fig:Quantization_Sensitivity_acc_drop_D}]{
        \includegraphics[width=0.23\textwidth]{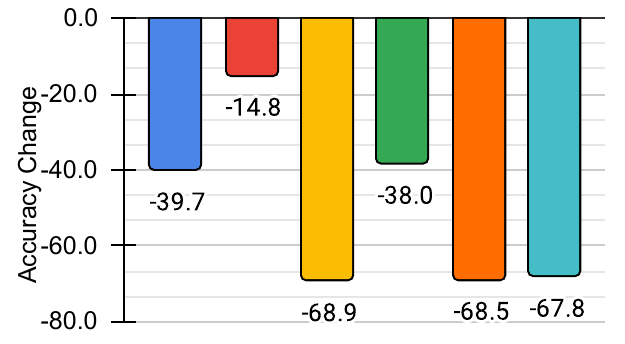}
    }
    
    \caption{
    Quantization sensitivity analysis of DeiT-Small transformer blocks using CLR-RQViT at different bit-widths. Each value shows the accuracy change when a target layer is quantized at a specific precision, with all other layers fixed at 4-bit precision, compared to the fully 4-bit quantized model.}
    \label{fig:Quantization_Sensitivity_acc_drop}
\end{figure}

We investigate the quantization sensitivity of individual transformer layers in ViTs using a PTQ approach across various bit settings. First, we fix the baseline quantization bits for all transformer layers. Then, we iteratively select a target layer and adjust its quantization bits across all transformer blocks, while keeping the other layers at their baseline precision. This process allows us to measure the impact on model performance and determine each layer's sensitivity to quantization at different precision levels. For instance, with a 4/4-bit baseline, we progressively adjust the target layer’s precision to 2/2, 3/3, 5/5, and 6/6 bits to identify sensitivity patterns. Experimental results on DeiT-S with ImageNet-1K, shown in Fig.~\ref{fig:Quantization_Sensitivity_acc_drop}, demonstrate how changes in precision affect network performance.

Unlike \cite{Q-ViT_2_Q_sensitivity} and \cite{Quantization_variation}, which identified the FC2 layer and the entire MHSA module, respectively, as highly sensitive to quantization, our findings reveal that quantization sensitivity varies across individual layers and is influenced by precision. For instance, increasing the precision of the MatMul2 operation from 4/4 to 5/5 or 6/6 bits had minimal impact, while reducing it to 3/3 bits resulted in significant performance degradation. In contrast, increasing the precision of the FC2 layer notably improved model performance compared to other layers. However, the quantization sensitivity score for all layers quantized at 2-bit precision showed a sharp decline in performance, primarily due to the limitations of the PTQ method in handling very low-bit quantization.

To evaluate the quantization sensitivity of each layer, we propose a \textbf{sensitivity score} that quantifies the model's responsiveness to precision changes across different bit widths, \(\mathcal{B} = \{b_1, b_2, \dots, b_{|\mathcal{B}|}\}\). Specifically, we calculate the performance change for each layer by comparing the loss of two configurations: (1) a baseline model where all quantized layers (\(U\)) are set to a uniform precision \(b'\), with \(b' \in \mathcal{B}\), and (2) a perturbed model where the target layer (\(T\), where \(T \in U\)) is quantized at \(b \in \mathcal{B}\), while all other layers (\(U \setminus T\)) remain at the baseline precision \(b'\). The performance change, typically negative for higher bit-width quantization, is calculated as the difference in loss between the baseline and perturbed configurations. To mitigate distortion from negative values, we adjust the performance change by adding the absolute value of the minimum observed change across all layers. Finally, the quantization sensitivity score is normalized as the ratio of the adjusted performance change for a given layer to the total adjusted change across all layers:  
\begin{align}
    \Delta \mathcal{L}_{T_{b}} &= \mathcal{L}_{{(U \setminus T)}_{b'}, T_b} - \mathcal{L}_{U_{b'}}, \label{eq:loss_change} \\
    \Delta \mathcal{L}_{T_{b}}^{+} &= \Delta \mathcal{L}_{T_{b}} + \left|\min_{u \in U, \, b \in \mathcal{B}} \Delta \mathcal{L}_{u_{b}}\right|, \label{eq:adjusted_loss} \\
    \Lambda_{T_b} &= \frac{\Delta \mathcal{L}_{T_{b}}^{+}}{\sum_{b \in \mathcal{B}} \sum_{u \in U_b} \Delta \mathcal{L}_{u_{b}}^{+}}. \label{eq:sensitivity_score}
\end{align}

\begin{figure}[t]
    \centering
    \subfloat[\label{fig:Quantization_Sensitivity_CRL_QViT}]{
        \includegraphics[width=0.23\textwidth]{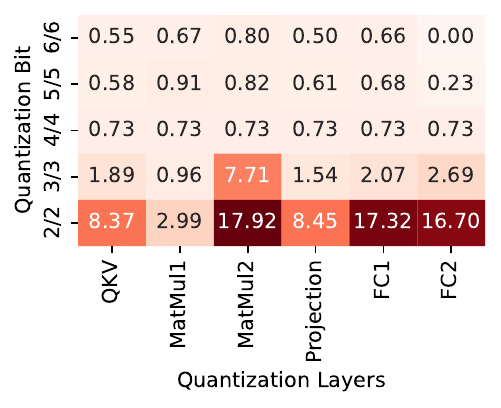}
    }%
    \hfill
    \subfloat[\label{fig:Quantization_Sensitivity_CRL_AdaLog}]{
        \includegraphics[width=0.23\textwidth]{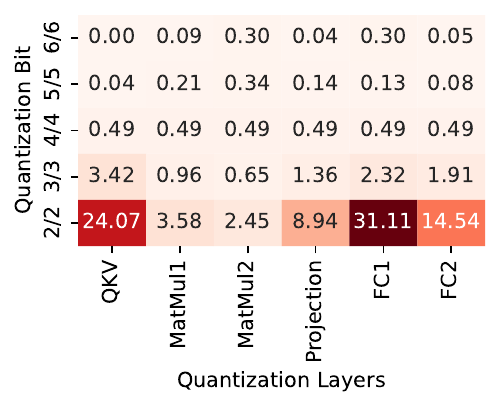}
    }%
    \caption{Quantization Sensitivity Score of DeiT-Small based on the (a) RQViT method and (b) AQViT method, respectively. Each value represents the layer's robustness to quantization as a percentage, with higher values indicating greater sensitivity to quantization error.}
    \label{fig:Quantization_Sensitivity_Score_DeiT_S}
\end{figure}


We estimated the quantization sensitivity score using a subset of 256 images from the ImageNet1K validation dataset, with minimal impact on results due to the small sample size. 
Detailed analysis of sample size effects is provided in Table~\ref{tab:Data Size & Component Ablation Study}. Figure~\ref{fig:Quantization_Sensitivity_Score_DeiT_S} shows the sensitivity scores of the DeiT-Small model using RQViT and AQViT for bit widths from 2/2 to 6/6, with a baseline of 4/4 bits. Layers with higher sensitivity are more prone to quantization loss, especially at 2/2 bits, reflecting greater error. 

\subsection{Mixed-Precision Bit Allocation Strategy}
Recent studies on CNNs~\cite{dong2019hawq, dong2020hawqv2, CLADO_deng2023mixed, MPQCO_chen2021towards} and ViTs~\cite{SQNR_tai2024mptq, SQNR_lee2022optimizing, Hessian_Mixed_languagexu2021mixed, learned_layer_tang2022mixed} have explored mixed-precision quantization to improve model compression and performance. 
It enables different layers to use different bit-widths, allowing efficient trade-offs between accuracy and resource usage.

For a given layer \( l \), there are \( |\mathcal{B}| \) possible bit-width choices from the set \(\mathcal{B}\), applicable to both weights and activations. These choices are represented as \( (b_w^{l}, b_a^{l}) \), where \( b_w^{l} \in \mathcal{B} \) and \( b_a^{l} \in \mathcal{B} \). The bit-width configuration for the entire architecture is \( \mathcal{S} = \{(b_w^{l}, b_a^{l})\}_{l=0}^{L} \), and the search space \( \mathcal{A} \) consists of all possible combinations of \( \mathcal{S} \). The goal of the MPQ method is to find the optimal bit-width configuration \( \mathcal{S}^{*} \in \mathcal{A} \) that maximizes validation performance, subject to resource constraints \( C \). This process is computationally intensive, requiring quantization, calibration, and evaluation for each candidate configuration. Despite the high cost, methods like AutoQ \cite{lou2020autoq} use this exhaustive search approach.

In this work, we introduce a mixed-precision bit selection strategy that uses pre-computed indicators, such as the layer importance score (\( \Omega \)) and quantization sensitivity score (\(\Lambda\)), to automatically assign bit-widths under constraints such as model size (\( \mathcal{C}_M \)) and bit operations (\(\mathcal{C}_{\text{BitOps}}\)). The objective is to maximize the optimization score (\( \Phi \)), which is the sum of each layer's importance score (\( \Omega_l \)) multiplied by its bit-width (\(\mathcal{B}_l\)), with an additional penalty term based on the product of the sensitivity score (\(\Lambda_l\)) and bit-width (\(\mathcal{B}_l\)). This approach eliminates the need for traditional iterative evaluations and is reformulated as an Integer Quadratic Programming (IQP) problem, as follows:
\begin{subequations}
\label{eq:Optimization_score_LIS_Quantization_Sensitivity(QS)}
\begin{align}
\Phi = \max &\left(\sum_{l=1}^{L} \left(\Omega_l \cdot \mathcal{B}_l - \Lambda_l \cdot \mathcal{B}_l\right) \right) \\
\text{s.t.} \quad & \sum_{l=1}^{L} M_l \leq \mathcal{C}_M  \\
& \sum_{l=1}^{L} \text{BitOps}_l(\mathcal{B}_l) \leq \mathcal{C}_{\text{BitOps}}
\end{align}
\end{subequations}
Here, \( M_l \) represents the model size for layer \( l \), and \( \text{BitOps}_l(\mathcal{B}_l) \) represents the bit operations for layer \( l \). 

In the absence of a quantization score term in (\ref{eq:Optimization_score_LIS_Quantization_Sensitivity(QS)}), the optimization function tends to assign higher bit-widths to layers with high importance scores while reducing the bit-widths of other layers to meet resource constraints. As shown in Table~\ref{tab:Data Size & Component Ablation Study}, this approach can lead to significant performance degradation, as lowering the bit-widths of quantization-sensitive layers negatively impacts overall model performance. To address this, the quantization score term is introduced to guide the objective function, ensuring that low bit-widths are not assigned to quantization-sensitive layers. This adjustment promotes a balanced allocation of bit-widths, mitigating the tendency to assign excessively high bits. A detailed analysis of this limitation is provided in Table~\ref{tab:Data Size & Component Ablation Study}.

For ViT with $L$ transformer block and each block consisting of multiple quantization layers $U$, where \(U=\{u_1, u_2, \ldots, u_{|U|}\}\). For a quantized layer \(u \in U\) in ViT block \(l\), we introduce a one-hot variable \(\alpha_{u}^{l} \in \{0, 1\}^{|\mathcal{B}|}\) such that \(\sum_{b=1}^{|\mathcal{B}|} \alpha_{u, b}^{l} = 1\) to represent the layer's bit-width decision. The MPQ problem in (\ref{eq:Optimization_score_LIS_Quantization_Sensitivity(QS)}) can be reformulated as the following:
\begin{subequations}
\begin{align}
\Phi = \max_{\alpha}  \sum_{l=1}^L \sum_{u \in U} &\Bigg( \Omega_u^l \odot \bigg( \sum_{j=1}^{|B|} \alpha_{u, j}^l b_j \bigg) 
- \sum_{j=1}^{|B|} \Lambda_{u, j} \alpha_{u, j}^l b_j \Bigg), \label{eq:Mixed_precision_bit_aallocation_LIS_QS} \\
\text{s.t.} \quad 
\alpha_{u, j}^l \in &\{0, 1\}, \quad \sum_{j=1}^{|B|} \alpha_{u, j}^l = 1, \quad \forall l, u, \label{eq:binary_constraint} \\
\sum_{l=1}^L \sum_{u \in U} &\sum_{j=1}^{|B|} \alpha_{u, j}^l |w_u^l| b_j \leq \mathcal{C}_M, \label{eq:memory_constraint} \\
\sum_{l=1}^L \sum_{u \in U} &\sum_{j=1}^{|B|} \text{MAC}_u^l \alpha_{u, j}^l b_j^2 \leq \mathcal{C}_{\text{BitOps}}, \label{eq:bitops_constraint} \\
l \in \{1, \dots, L\},  u &\in \{u_1, \dots, u_{|U|}\},  B \in \{b_1, \dots, b_{|B|}\}. \label{eq:indices}
\end{align}
\end{subequations}
Here, $\text{MAC}$ represents the multiply-and-accumulate operation of the layer, and $w$ represents weight of the the layer.

\section{Experiments}
\label{sec:result}

\subsection{Experimental Setup}



All pretrained weights in this work are sourced from the Timm library. For the layer importance score, we adopt the backbone architecture from~\cite{chefer2021LRP}, and for the quantization sensitivity score, we use the architectures of RQViT and AQViT. The mixed-precision bit allocation problem is solved using the CVXPY package~\cite{diamond2016cvxpy}.

We evaluate Mix-QViT in both PTQ and QAT settings. For PTQ, we assess three tasks: image classification on ImageNet1K~\cite{krizhevsky2012imagenet} (calibrated with 32 random training images), and object detection and instance segmentation on COCO~\cite{lin2014microsoft_COCO} (calibrated with a single sample). Percentile-based methods and clipped channel-wise scale reparameterization are applied to post-LayerNorm activations in all blocks. For QAT, we evaluate ImageNet1K, assigning optimal bit-widths via our mixed-precision strategy, followed by fine-tuning on the full dataset.

To ensure a fair comparison with the fixed-bit precision framework, our mixed-precision models are designed to maintain the same overall model size and bit-operation counts as their fixed-precision counterparts.

\subsection{Post Training Quantization}
\subsubsection{Results on ImageNet1K}
To demonstrate the effectiveness of our fixed-precision (RQViT and AQViT) and mixed-precision frameworks in image classification, we conducted extensive experiments on multiple vision transformer architectures, including ViT~\cite{image_16x16_dosovitskiy2021an_ViT}, DeiT~\cite{touvron2021training_DeiT}, and Swin~\cite{liu2021swin}. The quantization results under PTQ settings are summarized in Table~\ref{tab:ResultsImageNet_PTQ}, demonstrating significant performance gains over SOTA methods.

In the 3-bit fixed-precision setting, RepQ-ViT~\cite{li2023repq} suffers severe performance degradation, with classification accuracies dropping below 4\% across most architectures. In contrast, our RQViT achieves substantial improvements, reaching up to 65\% accuracy on the DeiT-Base architecture. Similarly, AQViT outperforms its counterpart, AdaLog~\cite{AdaLog_wu2025}, in the 3-bit configuration, achieving gains ranging from 1\% to 13\% across different architectures. In the 4-bit fixed-precision setting, both RQViT and AQViT continues to outperform, achieving 2–3\% higher accuracy compared to RepQ-ViT~\cite{li2023repq} and AdaLog~\cite{AdaLog_wu2025}, respectively. These results highlight the robustness of our fixed-precision frameworks across low-bit quantization scenarios.

The introduction of our mixed-precision strategy further enhances the performance. Mixed-precision applied to both RQViT (refered to as Mix-RQViT) and AQViT (referred to as Mix-AQViT) consistently outperforms their fixed-precision counterparts across all architectures. For 3-bit mixed-quantization, Mix-RQViT achieves improvements ranging from 6\% to 22\% over fixed-precision RQViT, with similar trend observed for Mix-AQViT. In the 4-bit quantization setting, while the performance gap between mixed- and fixed-precision is smaller, it remains significant, with Mix-RQViT and Mix-AQViT, delivering 2-4\% higher accuracy than their fixed-precision versions. 

At 6-bit quantization, both fixed- and mixed-precision frameworks achieve accuracies nearly on par with the full-precision baseline. For instance, Mix-AQViT achieves accuracies of 81.63\% and 83.27\% on the DeiT-B and Swin-B architectures, respectively. These results represent minimal accuracy drops of just 0.17\% and 0.09\% compared to full-precision models, underscoring the efficiency and robustness of our proposed approach.

\begin{figure}[t]
\centering
  \includegraphics[width=0.8\linewidth]{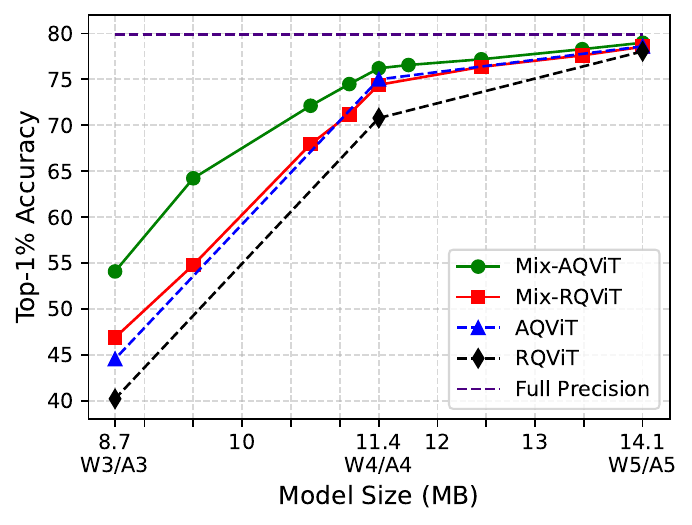}
  \caption{Model size vs. accuracy trade-off curves for DeiT-S under various fixed- and mixed-precision frameworks.}
  \label{fig:model_size_vs_acc_trade-off_DeiT-Small}
\end{figure}

Furthermore, Fig.~\ref{fig:model_size_vs_acc_trade-off_DeiT-Small} highlights the trade-off between model size and accuracy for DeiT-S. The model's accuracy improves as the model size increases, with more significant gains observed at smaller model sizes, such as those corresponding to 3-bit and 4-bit precision. For example, at a model size of 11.4 MB (4-bit precision), both of our mixed-precision frameworks demonstrate substantial improvements over their fixed-precision counterparts. Additionally, with a modest 9\% increase in model size (approximately 1 MB), Mix-AQViT and Mix-RQViT achieve further accuracy gains of 1\% and 1.95\%, respectively.

\begin{table*}
\caption{Quantization results for image classification on the ImageNet1K~\cite{krizhevsky2012imagenet} dataset. Each value represents the Top-1 accuracy (\%) obtained by quantizing the respective model. Here, ``Prec. (W/A)'' specifies the quantization bit precision for weights (W) and activations (A), respectively. ``CRL'' represents channel-wise Clipped Reparameterization of LayerNorm activation. ``MP'' denotes mixed precision, while ``$*$'' indicates that the model size and bit operations are equivalent to those of a fixed-bit quantized model. ``$\dagger$'' indicates replicated results that may not exactly match those reported in the original study. Bold values represent the best performance.}

  \label{tab:ResultsImageNet_PTQ}
  \centering
  \begin{tabular}{|l||c|c|c|c|c|c|c|r|}
   \hline
    Method&Prec.(W/A)&ViT-S&ViT-B&DeiT-T&DeiT-S&DeiT-B&Swin-S&Swin-B\\
    \hline \hline
    Full-Precision&32/32&81.39&84.54&72.21&79.85&81.80&83.23&85.27 \\
    \hline \hline
    PTQ4ViT~\cite{yuan2022ptq4vit}&3/3&0.10&0.10&3.50&0.10&31.06&28.69&20.13 \\
    RepQ-ViT~\cite{li2023repq}$^\dagger$&3/3&0.10&0.10&0.10&3.27&7.57&1.37&1.07\\
    AdaLog~\cite{AdaLog_wu2025}$^\dagger$&3/3& 12.63& 29.42& 25.70& 22.82& 55.90& 58.12& 61.54 \\
    RQViT (\!\!\cite{li2023repq}+CRL)&3/3&6.63&2.72&15.30&40.22&64.54&20.42&23.32 \\
    AQViT (\!\!\cite{AdaLog_wu2025}+CRL) &3/3&\textbf{13.08}&\textbf{47.55}&\textbf{27.54}&\textbf{44.56}&\textbf{65.30}&\textbf{65.08}&\textbf{69.11} \\
    \cline{2-9}
    Mix-RQViT (ours)$^*$ &MP3/MP3&14.88&23.36&28.64&{46.89}&{65.91}&32.18&35.41\\
    
    Mix-AdaLog (\!\!\cite{AdaLog_wu2025}+MP)$^*$ &MP3/MP3&18.39&36.41&32.37&29.14&59.88&{66.23}&{67.55} \\
    Mix-AQViT (ours)$^*$ &MP3/MP3&\textbf{21.44}&\textbf{53.36}&\textbf{34.83}&\textbf{54.08}&\textbf{68.32}&\textbf{69.66}&\textbf{71.18}\\
    \hline \hline
    FQ-ViT~\cite{lin2022fq_vit}&4/4&0.10&0.10&0.10&0.10&0.10&0.10&0.10 \\
    PTQ4ViT~\cite{yuan2022ptq4vit}&4/4&42.57&30.69&36.96&34.08&64.39&76.09&74.02 \\
    APQ-ViT~\cite{ding2022APQViTtowards}&4/4&47.95&41.41&47.94&43.55&67.48&77.15&76.48 \\
    RepQ-ViT~\cite{li2023repq}&4/4&65.05&68.48&57.43&69.03&75.61&79.45&78.32\\
    AdaLog~\cite{AdaLog_wu2025}&4/4&72.75&79.68&63.52&72.06&78.03&80.77&82.47 \\
    RQViT (\!\!\cite{li2023repq}+CRL)&4/4&68.25&73.54&59.06&70.78&77.40&80.55&80.04 \\
    AQViT (\!\!\cite{AdaLog_wu2025}+CRL) &4/4&\textbf{74.06}&\textbf{81.10}&\textbf{64.03}&\textbf{74.99}&\textbf{79.51}&\textbf{81.29}&\textbf{83.27} \\
    \cline{2-9}
    Mix-RQViT (ours)$^*$ &MP4/MP4&{72.34}&{76.39}&{63.53}&{74.40}&{78.91}&{81.11}&{82.05} \\
    
    Mix-AdaLog (\!\!\cite{AdaLog_wu2025}+MP)$^*$ &MP4/MP4&73.57&80.95&63.87&{74.73}&{79.15}&{81.42}&{82.87} \\
    Mix-AQViT (ours)$^*$ &MP4/MP4&\textbf{75.61}&\textbf{82.56}&\textbf{65.24}&\textbf{76.20}&\textbf{80.33}&\textbf{82.16}&\textbf{84.14} \\
    \hline \hline
    FQ-ViT~\cite{lin2022fq_vit}&6/6&4.26&0.10&58.66&45.51&64.63&66.50&52.09 \\
    PSAQ-ViT~\cite{li2022patch}&6/6&37.19&41.52&57.58&63.61&67.95&72.86&76.44 \\
    Ranking~\cite{liu2021post_Rank_Aware_PTQ}&6/6&-&75.26&-&74.58&77.02&-&- \\
    PTQ4ViT~\cite{yuan2022ptq4vit}&6/6&78.63&81.65&69.68&76.28&80.25&82.38&84.01 \\
    APQ-ViT~\cite{ding2022APQViTtowards}&6/6&79.10&82.21&70.49&77.76&80.42&82.67&84.18 \\
    RepQ-ViT~\cite{li2023repq}&6/6&80.43&83.62&70.76&78.90&81.27&82.79&84.57 \\
    AdaLog~\cite{AdaLog_wu2025}&6/6&80.91&84.80&71.38&79.39&81.55&{83.19}&85.09 \\
    RQViT (\!\!\cite{li2023repq}+CRL)&6/6&80.52&83.83&70.96&79.00&81.39&82.77&84.63\\
    AQViT (\!\!\cite{AdaLog_wu2025}+CRL) &6/6&\textbf{80.94}&\textbf{84.81}&\textbf{71.42}&\textbf{79.42}&\textbf{81.57}&\textbf{83.25}&\textbf{85.12} \\
    \cline{2-9}
    Mix-RQViT (ours)$^*$ &MP6/MP6&{80.97}&{84.12}&{71.49}&{79.37}&{81.53}&{82.91}&{84.86} \\
    Mix-AdaLog (\!\!\cite{AdaLog_wu2025}+MP)$^*$&MP6/MP6&{80.93}&84.47&{71.66}&{79.54}&{81.61}&{83.11}&{85.15} \\
    Mix-AQViT (ours)$^*$ &MP6/MP6&\textbf{80.99}&\textbf{84.86}&\textbf{71.70}&\textbf{79.62}&\textbf{81.63}&\textbf{83.27}&\textbf{85.18}  \\
    \hline 
  \end{tabular}
\end{table*}

\begin{table*}[t]
  \small
  \caption{Quantization results on object detection and instance segmentation on the COCO~\cite{lin2014microsoft_COCO} dataset. Here, $AP^{box}$ is the box average precision for object detection, and $AP^{mask}$ is the mask average precision for instance segmentation. 
  ``Prec. (W/A)" indicates the quantization bit-precision for weights and activations as (W) and (A) bits, respectively. ``CRL'' represents channel-wise Clipped Reparameterization of LayerNorm activation. ``MP'' denotes mixed precision, while ``$*$'' indicates that the model size and bit operations are equivalent to those of a fixed-bit quantized model. ``$\dagger$'' indicates replicated results that may not exactly match those reported in the original study. Bold values represent the best performance.}
  \label{tab:ResultsCOCO}
  \centering
  \begin{tabular}{|l||c|cc|cc|cc|cr|}
    \hline 
    \multirow{3}{*}{Method} & \multirow{3}{*}{Prec. (W/A)} & \multicolumn{4}{c|}{Mask R-CNN} & \multicolumn{4}{c|}{Cascade Mask R-CNN} \\
    \cline{3-6} \cline{7-10}
    & & \multicolumn{2}{c}{w.Swin-T} & \multicolumn{2}{c|}{w.Swin-S} & \multicolumn{2}{c}{w.Swin-T} & \multicolumn{2}{c|}{w.Swin-S} \\
    \cline{3-10}
    & & AP$^{box}$ & AP$^{mask}$ & AP$^{box}$ & AP$^{mask}$ & AP$^{box}$ & AP$^{mask}$ & AP$^{box}$ & AP$^{mask}$ \\
    \hline \hline
    Full-Precision & 32/32 & 46.0 & 41.6 & 48.5 & 43.3 & 50.4 & 43.7 & 51.9 & 45.0 \\
    \hline \hline
    RepQ-ViT~\cite{li2023repq}$^\dagger$ & 3/3 & 0.5 & 0.5 & 1.9 & 1.3 & 0.7 & 0.7 & 1.3 & 1.2 \\
    AdaLog~\cite{AdaLog_wu2025}$^\dagger$ & 3/3 & 12.6 & 11.4 & 21.0 & 19.4 & 20.8 & 15.6 & 25.6 & 19.7 \\
    RQViT (\!\!\cite{li2023repq}+CRL) & 3/3 & 2.8 & 2.2 & 11.8 & 10.1 & 4.1 & 3.3 & 12.2 & 12.9 \\
    AQViT (\!\!\cite{AdaLog_wu2025}+CRL) & 3/3 & \textbf{21.1} & \textbf{20.3} & \textbf{30.7} & \textbf{24.6} & \textbf{30.9} & \textbf{26.2} & \textbf{32.3} & \textbf{27.4} \\
    \cline{2-10}
    Mix-RQViT (ours)$^*$  & MP3/MP3 & {4.6} & {4.2} & {13.2} & 12.4 & 8.3 & {8.0} & {15.2} & {13.5} \\
    Mix-AdaLog (\!\!\cite{AdaLog_wu2025}+MP)$^*$ & MP3/MP3 & {23.8} & {22.2} & {30.5} & 26.1 & {32.2} & {28.1} & {35.3} & {29.4} \\
    Mix-AQViT (ours)$^*$ & MP3/MP3 & \textbf{31.6} & \textbf{29.1} & \textbf{35.4} & \textbf{32.3} & \textbf{36.2} & \textbf{32.8} & \textbf{40.7} & \textbf{33.3} \\
    \hline \hline
    PTQ4ViT~\cite{yuan2022ptq4vit} & 4/4 & 6.9 & 7.0 & 26.7 & 26.6 & 14.7 & 13.5 & 0.5 & 0.5 \\
    APQ-ViT~\cite{ding2022APQViTtowards} & 4/4 & 23.7 & 22.6 & 44.7 & 40.1 & 27.2 & 24.4 & 47.7 & 41.1 \\
    RepQ-ViT~\cite{li2023repq} & 4/4 & 36.1 & 36.0 & 44.2 & 40.2 & 47.0 & 41.4 & 49.3 & 43.1 \\
    AdaLog~\cite{AdaLog_wu2025} & 4/4 & 39.1 & 37.7 & 44.3 & 41.2 & 48.2 & 42.3 & 50.6 & 44.0 \\
    TSPTQ-ViT~\cite{tai2023tsptq} & 4/4 & \textbf{42.9} & 39.3 & 45.0 & 40.7 & 47.8 & 41.6 & 48.8 & 42.5 \\
    RQViT (\!\!\cite{li2023repq}+CRL) & 4/4 & 38.6 & 37.4 & 44.3 & 40.6 & 47.4 & 41.5 & 49.5 & 43.4 \\
    AQViT (\!\!\cite{AdaLog_wu2025}+CRL) & 4/4 & 41.8 & \textbf{39.6} & \textbf{45.4} & \textbf{41.8} & \textbf{48.8} & \textbf{42.5} & \textbf{50.9} & \textbf{44.2} \\
    \cline{2-10}
    MPTQ-ViT~\cite{SQNR_tai2024mptq} & MP4/MP4 & {44.2} & {40.2} & 47.3 & \textbf{42.7} & {49.2} & {42.7} & 50.8 & 44.2 \\
    Mix-RQViT (ours)$^*$  & MP4/MP4 & {43.5} & {39.4} & {47.3} & 42.3 & 48.7 & {42.6} & {50.9} & \textbf{44.6} \\
    Mix-AdaLog (\!\!\cite{AdaLog_wu2025}+MP)$^*$ & MP4/MP4 & {43.8} & {40.3} & {47.4} & 42.5 & {49.4} & {42.8} & {51.0} & \textbf{44.6} \\
    Mix-AQViT (ours)$^*$ & MP4/MP4 & {44.8} & \textbf{40.6} & \textbf{47.6} & \textbf{42.7} & \textbf{49.6} & \textbf{43.0} & \textbf{51.3} & \textbf{44.6} \\
    \hline \hline
    PTQ4ViT~\cite{yuan2022ptq4vit} & 6/6 & 5.8 & 6.8 & 6.5 & 6.6 & 14.7 & 13.6 & 12.5 & 10.8 \\
    APQ-ViT~\cite{ding2022APQViTtowards} & 6/6 & 45.4 & 41.2 & 47.9 & 42.9 & 48.6 & 42.5 & 50.5 & 43.9 \\
    RepQ-ViT~\cite{li2023repq} & 6/6 & 45.1 & 41.2 & 47.8 & 43.0 & 50.0& 43.5& 51.4 & 44.6 \\
    AdaLog~\cite{AdaLog_wu2025} & 6/6 & 45.4 & 41.3 & 48.0 & \textbf{43.2} & 50.1 & \textbf{43.6} & 51.7 & \textbf{44.8} \\
    TSPTQ-ViT~\cite{tai2023tsptq} & 6/6 & 45.8 & \textbf{41.4} & \textbf{48.3} & \textbf{43.2} & \textbf{50.2} & 43.5 & \textbf{51.8} & \textbf{44.8} \\
    RQViT (\!\!\cite{li2023repq}+CRL) & 6/6 & 45.3 & 41.2 & 47.9 & 42.9 & 50.0 & 43.5 & 51.5 & 44.6 \\
    AQViT (\!\!\cite{AdaLog_wu2025}+CRL) & 6/6 & 45.8 & \textbf{41.4} & {48.1} & \textbf{43.2} & \textbf{50.2} & \textbf{43.6} & \textbf{51.8} & \textbf{44.8} \\
    \cline{2-10}
    MPTQ-ViT~\cite{SQNR_tai2024mptq} & MP6/MP6 & \textbf{45.9} & 41.4 & \textbf{48.3} & 43.1 & \textbf{50.2} & \textbf{43.6} & \textbf{51.8} &44.8 \\
    Mix-RQViT (ours)$^*$ & MP6/MP6 & 45.8 & 41.4 & 48.0 & \textbf{43.2} & \textbf{50.2} & \textbf{43.6} & \textbf{51.8} & 44.8 \\
    Mix-AdaLog (\!\!\cite{AdaLog_wu2025}+MP)$^*$ & MP6/MP6 & \textbf{45.9} & 41.4 & \textbf{48.3} & \textbf{43.2} & \textbf{50.2} & \textbf{43.6} & \textbf{51.8} & 44.8 \\
    Mix-AQViT (ours)$^*$ & MP6/MP6 & \textbf{45.9} & \textbf{41.5} & \textbf{48.3} & \textbf{43.2} & \textbf{50.2} & \textbf{43.6} & \textbf{51.8} & \textbf{44.9} \\
    \hline 
  \end{tabular}
  
\end{table*}

\subsubsection{Results on COCO}

To further assess the effectiveness of Mix-QViT, we conducted experiments on object detection and instance segmentation tasks using the Mask R-CNN~\cite{he2017maskRCNN} and Cascade Mask R-CNN~\cite{cai2018cascade_Mask_RCNN} detectors, with quantized Swin transformers as the backbone. The results, summarized in Table~\ref{tab:ResultsCOCO}, demonstrate the robustness of our approach. In the 3-bit setting, while RepQ-ViT~\cite{li2023repq} suffers from severe performance degradation and AdaLog~\cite{AdaLog_wu2025} shows moderate improvement but remains suboptimal, our fixed-bit frameworks outperform their counterparts by significant margins, with AQViT achieving particularly strong results. When mixed-precision is introduced, our Mix-AQViT achieves even greater performance. For 4-bit quantization, both our fixed- and mixed-precision frameworks continue to deliver superior results, consistently outperforming state-of-the-art methods. At 6-bit quantization, both Mix-RQViT and Mix-AQViT surpass all other methods and achieve results nearly identical from those of the full-precision models.

\subsection{Quantization-Aware Training}

\begin{table}[ht]
\caption{Quantization-Aware Training results on the ImageNet1K validation dataset. Each value represents the Top-1 accuracy (\%) for quantized models. ``Prec. (W/A)'' specifies the quantization bit precision for weights (W) and activations (A). ``MP'' denotes mixed precision, ``FP'' represents full precision, and ``$*$'' indicates that the model size and bit operations are equivalent to those of a fixed-bit quantized model. Bold values highlight the best performance.}
\label{tab:ResultsImageNet_QAT}
\centering
\begin{tabular}{|l||c|c|c|c|c|r|}
\hline
\multirow{2}{*}{Model} & \multirow{2}{*}{FP} & \multicolumn{1}{c|}{LSQ} & \multicolumn{1}{c|}{Q-ViT} & \multicolumn{1}{c|}{OFQ} & \multicolumn{1}{c|}{Mix-} & \multicolumn{1}{c|}{Mix-} \\
& & \cite{esser2019learned} & \cite{li2022QViT} & \cite{liu2023oscillation} &LSQ$^*$ & OFQ$^*$\\
\hline \hline
DeiT-T & 72.21 & 54.45 & 50.37 & 64.33 & 64.19 & \textbf{67.87} \\
DeiT-S & 79.85 & 68.00 & 72.10 & 75.72 & 73.88 & \textbf{76.39} \\
DeiT-B & 81.80 & 70.30 & 74.20 & - & 76.58 & \textbf{78.26} \\
Swin-T & 81.20 & 70.40 & 74.70 & 78.52 & 75.13 & \textbf{78.71} \\
Swin-S & 83.23 & 72.40 & 76.90 & - & 79.49 & \textbf{81.23} \\
\hline
\end{tabular}
\end{table}

When access to the full training dataset and pipeline is available, integrating quantization-aware training (QAT) with mixed-precision quantization (MPQ) offers an opportunity to compress the network while maintaining high performance. To demonstrate the effectiveness of our MPQ strategy in QAT settings, we conducted experiments on ViT, DeiT, and Swin models using mixed-precision bit-assignment configurations.

Specifically, we evaluated a 2-bit mixed-precision setting by quantizing the weights and activations of all layers using candidate bits \(\mathcal{B} = \{1, 2, 3\}\), while preserving 8-bit precision for the patch embedding and classifier head. For bit allocation, we used layer importance scores as the guiding metric and imposed an optimization constraint to ensure that the total model size and bit operations remained equivalent to those of a 2-bit fixed-precision model. Table~\ref{tab:ResultsImageNet_QAT} presents the QAT results on the ImageNet1K validation dataset.

To benchmark our approach, we adopted LSQ~\cite{esser2019learned} and OFQ~\cite{liu2023oscillation} as baseline QAT methods with our mixed-precision framework.
Incorporating LSQ into Mix-QViT (referred to as Mix-LSQ), our mixed-precision method achieves a significant improvement in model accuracy, with an average increase of up to 6\% across all transformer models compared to fixed-bit quantization. Similarly, when combined with OFQ (referred to as Mix-OFQ), our framework achieves even greater performance gains. Mix-OFQ delivers near full-precision accuracy, such as 81.23\% on Swin-S, while maintaining computational efficiency. 

\begin{table} [t]
\caption{Ablation studies of different quantizers for post-LayerNorm activation. Here, W/A represents weights and activation bit. Each value represents the Top-1 accuracy (\%) for quantized models.
``CRL'' represents channel-wise (CW) Clipped Reparameterization of LayerNorm activation. Bold values highlight the best performance.}
  \label{tab:Ablation_Study_LN}
  \small
  \centering
  \begin{tabular}{|l|c||c|r|}
    \hline 
    \multirow{2}{*}{Model}&\multirow{2}{*}{Method}&\multicolumn{2}{c|}{Prec. (W/A)}\\
    \cline{3-4}
    &&3/3& 4/4 \\
    \hline \hline
    &Full-Precision&79.85&79.85\\
    \cline{2-4}
    \noalign{\vskip 1pt} 
    &Layer-wise Quant&0.10&33.17\\
    &CW Quant&18.33&70.28\\
DeiT-S&RepQ-ViT~\cite{li2023repq}&6.27&69.03\\
    &AdaLog~\cite{AdaLog_wu2025}&22.82&72.06\\
    &RQViT (\!\!\cite{li2023repq}+CRL)&\textbf{40.22}&7\textbf{0.78}\\
    &AQViT (\!\!\cite{AdaLog_wu2025}+CRL)&\textbf{44.56}&\textbf{74.99}\\
    \hline \hline
    &Full-Precision&83.23&83.23\\
    \cline{2-4}
    \noalign{\vskip 1pt} 
    &Layer-wise Quant&0.10&57.63\\
    &CW Quant&8.57&80.52\\
Swin-S&RepQ-ViT~\cite{li2023repq}&1.37&79.45\\
    &AdaLog~\cite{AdaLog_wu2025}&58.12&80.77\\
    &RQViT (\!\!\cite{li2023repq}+CRL)&\textbf{20.42}&\textbf{80.55}\\
    &AQViT (\!\!\cite{AdaLog_wu2025}+CRL)&\textbf{65.08}&\textbf{81.29}\\
    \hline 
  \end{tabular}
\end{table}

\begin{figure*}[t]
    \centering
    \subfloat[\label{fig:MPB_Allocation_a}]{
        \includegraphics[width=0.31\textwidth]{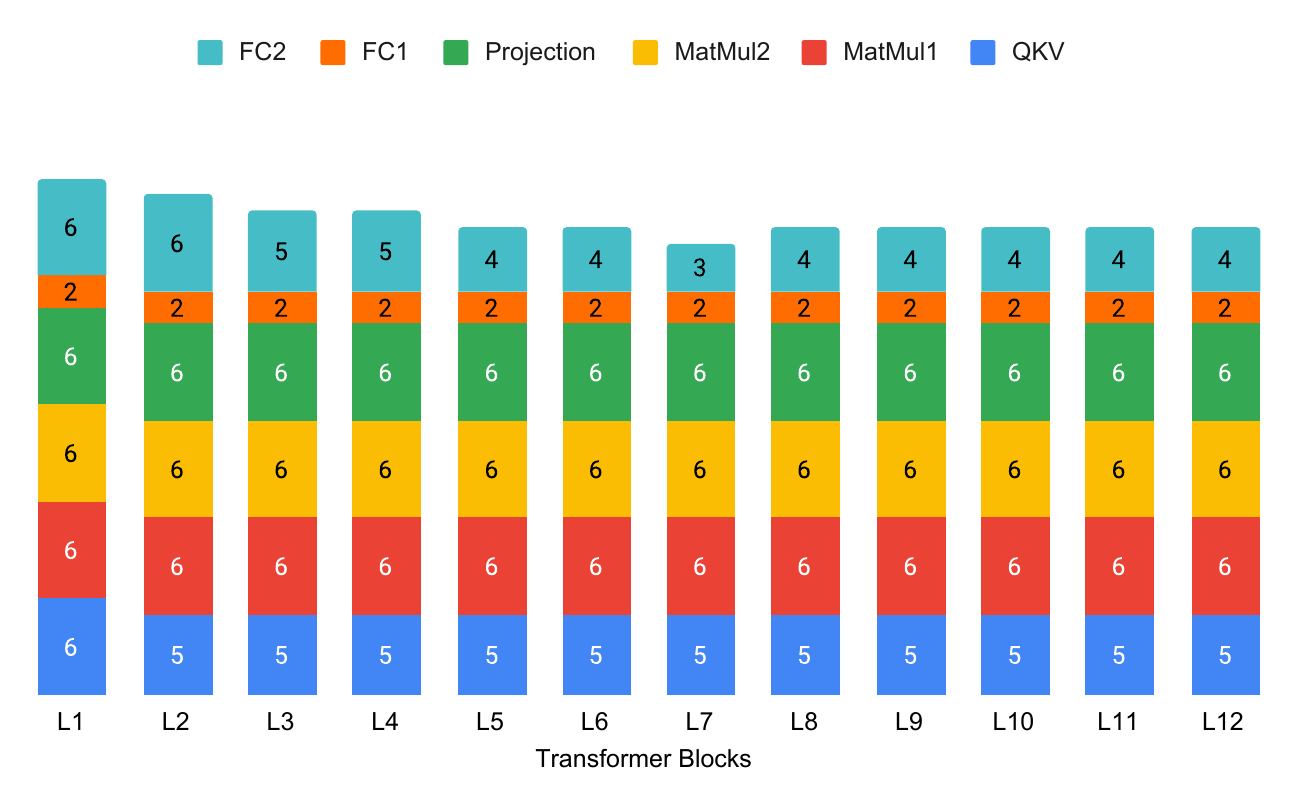}
    } \hfill
    \subfloat[\label{fig:MPB_Allocation_b}]{
        \includegraphics[width=0.31\textwidth]{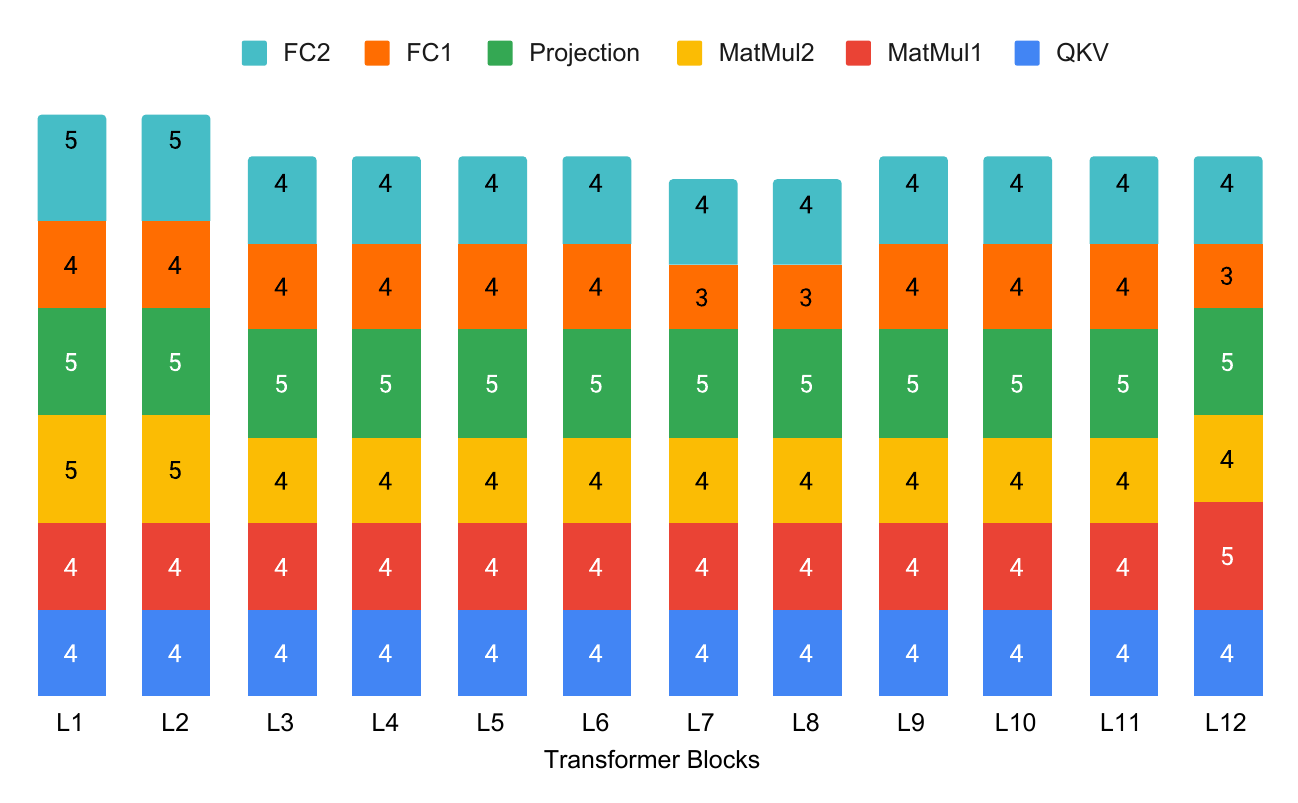}
    } \hfill
    \subfloat[\label{fig:MPB_Allocation_c}]{
        \includegraphics[width=0.31\textwidth]{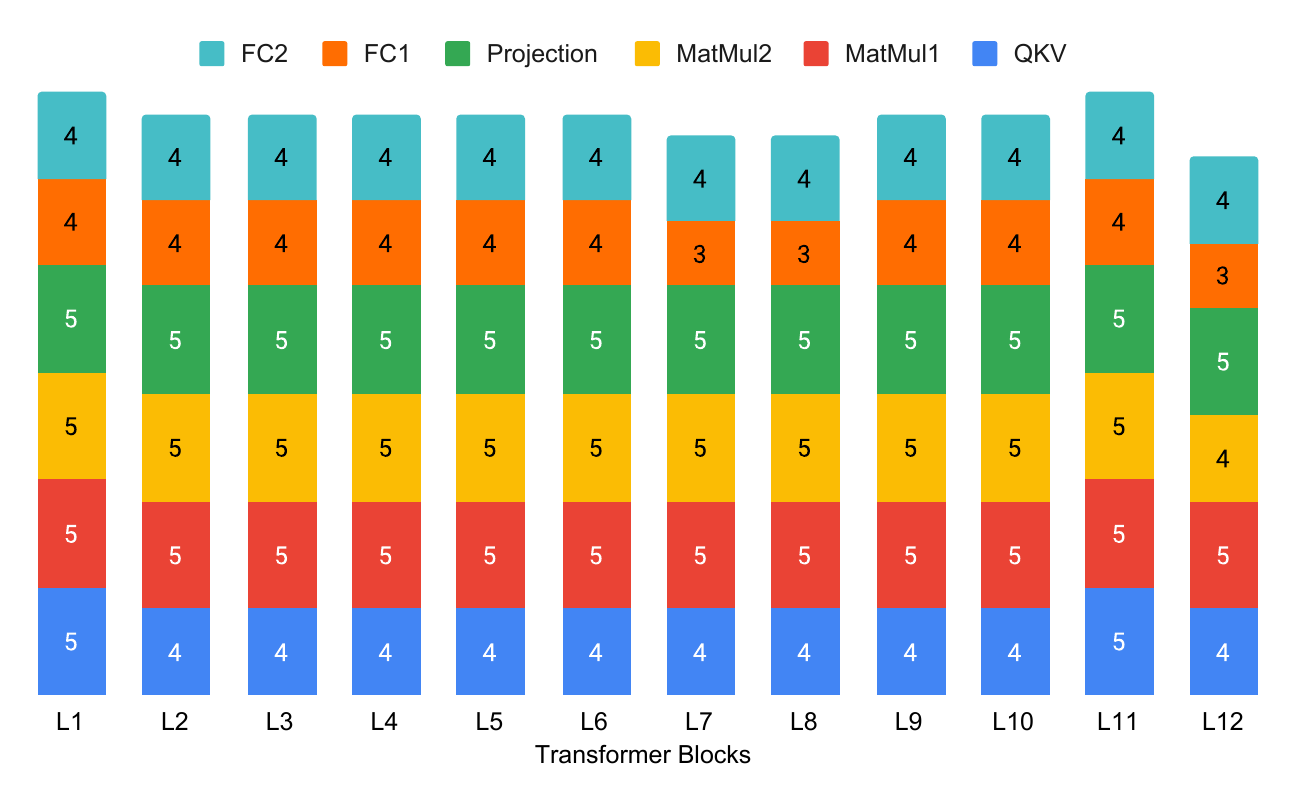}
    }
    \caption{
    Mixed-Precision Bit Allocation for DeiT-Small. 
    (a) Relies solely on the layer importance score, assigning higher precision bits to more important layers to maximize the optimization function in (Eq.~\ref{eq:Mixed_precision_bit_aallocation_LIS_QS}), while lower importance layers receive fewer bits. 
    (b-c) Combine the layer importance score with a quantization sensitivity score, which acts as a penalty term to discourage low-precision assignment in highly sensitive layers and to reduce unnecessary high-precision usage. (b) the sensitivity score is derived from RQViT, and (c) from AQViT. Candidate bit widths are \(\{2, 3, 4, 5, 6\}\).
    }
    \label{fig:MPB_Allocation}
\end{figure*}

\subsection{Ablation Study}
To verify the effectiveness of our proposed framework, we conducted two ablation studies: the impact of clipped channel-wise quantization for post-LayerNorm, summarized in Table~\ref{tab:Ablation_Study_LN}, and the role of input sample size on layer importance and quantization sensitivity scores, leading to mixed-precision bit allocation in PTQ, detailed in Table~\ref{tab:Data Size & Component Ablation Study}.

Table~\ref{tab:Ablation_Study_LN} highlights the effectiveness of various quantization strategies for post-LayerNorm activations on DeiT-S and Swin-S models under two bit precision settings: W3/A3 and W4/A4. Layer-wise quantization performs poorly, resulting in severe accuracy degradation, with accuracies dropping to just 0.10\% and 33.17\% on DeiT-S, and 0.10\% and 57.63\% on Swin-S. Channel-wise quantization significantly mitigates these issues, achieving 11.33\% and 70.28\% on DeiT-S, and 3.57\% and 80.52\% on Swin-S under W3/A3 and W4/A4, respectively. Our clipped channel-wise quantization (RQViT and AQViT) further enhances performance by effectively addressing outliers in post-LayerNorm activations during both calibration and inference stages. For instance, RQViT achieves 40.22\% (+21.89\%) for W3/A3 and 70.78\% (+0.5\%) for W4/A4 on DeiT-S, and 20.42\% (+11.85\%) for W3/A3 and 81.29\% (+0.03\%) for W4/A4 on Swin-S, outperforming standard channel-wise quantization. The improvements are even more pronounced with AQViT. The benefits of our method are particularly evident in challenging 3-bit quantization scenarios, where it consistently delivers significant and reliable performance improvements. The practical impact of clipped channel-wise quantization on model inference efficiency, including metrics such as throughput and calibration time, is presented in Table~\ref{tab:Efficiency_analysis_Data_Time_Throughput}.

\begin{table}[ht]
\caption{Ablation study on the impact of data sample size on the layer importance score (\( \Omega \)) and quantization score (\( \Lambda \)), as well as their roles in the mixed-precision quantization strategy. Each value represents Top-1\% accuracy, under mixed-precision (MP4/MP4) for weights and activations with bitwidth candidates \(\{2, 3, 4, 5, 6\}\). Quantization sensitivity scores (\({\Lambda}_1\)) and (\({\Lambda}_2\)) are derived from the RQViT and AQViT models, respectively.}
\label{tab:Data Size & Component Ablation Study}
\small
\centering
\resizebox{\columnwidth}{!}{%
\begin{tabular}{|ccc||cc|cr|}
    \hline
    \multicolumn{3}{|c||}{\# Image Samples } & \multicolumn{2}{c|}{DeiT-Small} & \multicolumn{2}{c|}{Swin-Small} \\
    \cline{1-3} \cline{4-5}  \cline{6-7} 
    \multirow{2}{*}{\( \Omega \)}& \multirow{2}{*}{\({\Lambda}_1\)} & \multirow{2}{*}{\({\Lambda}_2\)} & \multicolumn{1}{c}{Mix-} & \multicolumn{1}{c|}{Mix-} & \multicolumn{1}{c}{Mix-} & \multicolumn{1}{c|}{Mix-} \\ 
    &&&RQViT&AQViT&RQViT&AQViT\\
    \hline \hline
    50,000 & 50,000& \ding{53} & 74.62 & 73.13 & 81.57 & 80.13 \\
    50,000 & \ding{53} & 50,000 & 72.09 & 76.48 & 79.07 & 82.67 \\
    50,000 & 256 & \ding{53} & 74.51 & 72.89 & 81.52 & 79.85 \\
    50,000 & \ding{53} & 256 & 71.83 & 76.42 & 79.01 & 82.59 \\
    256 & 50,000 & \ding{53} & 74.43 & 72.66 & 81.16 & 79.30 \\
    256 & \ding{53} & 50,000 & 71.41 & 76.24 & 78.29 & 82.23 \\
    \hline \hline
    \textbf{256} & \textbf{256} & \ding{53} & \textbf{74.40} & 72.60 & \textbf{81.11} & 79.29 \\
    \textbf{256} & \ding{53} & \textbf{256} & 71.36 & \textbf{76.20} & 78.23 & \textbf{82.16} \\
    \hline \hline
    256 & \ding{53} & \ding{53} & 1.50 & 36.41 & 7.14 & 42.15 \\
    \ding{53} & 256 & \ding{53} & 0.37 & 1.37 & 1.03 & 3.21 \\
    \ding{53} & \ding{53} & 256 & 0.16 & 11.39 & 0.54 & 15.42 \\
    \hline 
\end{tabular}%
}
\end{table}

Figure~\ref{fig:MPB_Allocation} illustrates the mixed-precision bit allocations for DeiT-S with candidate bit widths \(\{2, 3, 4, 5, 6\}\). In Figure~\ref{fig:MPB_Allocation}(a), only the layer importance score (\(\Omega\)) is considered under the optimization constraints in (\ref{eq:Mixed_precision_bit_aallocation_LIS_QS}), which assigns 6-bit precision to the MatMul1, MatMul2, and Projection layers, and 2-bit precision to the FC1 layers. In contrast, Figures~\ref{fig:MPB_Allocation}(b) and \ref{fig:MPB_Allocation}(c) combine (\(\Omega\)) with a quantization sensitivity score (\(\Lambda\)), which penalizes using lower precision for quantization sensitive layers and reduces excessive use of high-precision bits. Specifically, Figure~\ref{fig:MPB_Allocation}(b) applies \(\Lambda_1\) derived from the RQViT framework, allowing only a few layers at 3-bit but avoiding 2-bit entirely and selectively using 5-bit for highly important layers. Figure~\ref{fig:MPB_Allocation}(c) adopts \(\Lambda_2\) from AQViT, resulting in a similar overall configuration.

Table~\ref{tab:Data Size & Component Ablation Study} presents an ablation study on the impact of data sample size when computing the layer importance score (\(\Omega\)) and quantization sensitivity score \((\Lambda\)), along with their individual contributions to mixed-precision quantization in DeiT-Small and Swin-Small. The layer importance score (\(\Omega\)), obtained via Layer-wise Relevance Propagation~\cite{chefer2021LRP}, remains the same across different quantization methods. In contrast, the quantization sensitivity score \(\Lambda\) depends on the underlying PTQ frameworks.

For the DeiT-S model, relying solely on \(\Omega\) in Mix-QViT drastically reduces accuracy to 1.50\%. Once \(\Lambda_1\) is included, as in Figure~\ref{fig:MPB_Allocation}(b), accuracy increases to 74.40\%. Using \(\Lambda_2\) in Mix-AQViT (Figure~\ref{fig:MPB_Allocation}(c)) achieves 76.20\%.
A similar pattern emerges for Swin-S, where ignoring \(\Lambda\) significantly degrades accuracy, yet including it not only recovers, but often surpasses, the performance of fixed-bit quantization. Furthermore, varying the data size (e.g., 50,000 versus 256 images) minimally affects outcomes (e.g., 74.62\% vs. 74.40\% in Mix-QViT on DeiT-S). Overall, these results confirm that combining \(\Omega\) with a framework-specific \(\Lambda\) yields more optimal mixed-precision allocations that balance accuracy and efficiency.


\begin{table}[ht]
\caption{
Comparison of data quantity, calibration time (Cali.) in minutes, and inference throughput (TP). Here, ``Prec. (W/A)'' specifies the quantization bit precision for weights (W) and activations (A), and ``$*$'' indicates mixed-precision models with model size and bit operations equivalent to a fixed-bit quantized model.
 ``CRL'' represents channel-wise Clipped Reparameterization of LayerNorm activation.}
\label{tab:Efficiency_analysis_Data_Time_Throughput}
\small
\centering
\resizebox{\columnwidth}{!}{%
\begin{tabular}{|l|c||c|c|c|r|}
\hline 
\multirow{2}{*}{Model} & \multirow{2}{*}{Method} & \multicolumn{2}{c|}{Prec. (W/A)} & \multicolumn{1}{c|}{Cali.} & \multicolumn{1}{c|}{TP} \\
\cline{3-4}
& & 3/3& 4/4& (min) & (img/s) \\
\hline \hline
\multirow{8}{*}{DeiT-S} 
& Full-Precision & 79.85& 79.85 & - & {551.8} \\
\cline{2-6}
& RepQ-ViT~\cite{li2023repq} &3.27& 69.03  & 1.8 & 497.2 \\
& AdaLog~\cite{AdaLog_wu2025} &22.82& 72.06  & 4.6 & 424.7 \\
& RQViT (\!\!\cite{li2023repq}+CRL)&40.22& 70.78  & 1.9 & 459.1 \\
&AQViT (\!\!\cite{AdaLog_wu2025}+CRL)&44.56& 74.99  & 4.6 & 420.0 \\
\cline{2-6}
& Mix-RQViT (ours)$^*$ &46.89& 74.40  & 2.0 & 437.2 \\
& Mix-AdaLog (\!\!\cite{AdaLog_wu2025}+MP)$^*$ &29.14& 74.73 &  4.6 & 424.4 \\
& Mix-AQViT (ours)$^*$ &54.08& 76.20 & 4.7 & 422.1 \\
\hline \hline
\multirow{8}{*}{Swin-S} 
& Full-Precision & 83.23 &83.23  & - & {510.8} \\
\cline{2-6}
& RepQ-ViT~\cite{li2023repq} &1.37& 79.45  & 3.5 & 285.8 \\
& AdaLog~\cite{AdaLog_wu2025} &58.12& 80.77 & 11.2 & 205.7 \\
&RQViT (\!\!\cite{li2023repq}+CRL)&20.42& 80.55  & 3.7 & 247.8 \\
&AQViT (\!\!\cite{AdaLog_wu2025}+CRL)&65.08& 81.29  & 11.2 & 203.3 \\
\cline{2-6}
& Mix-RQViT (ours)$^*$ &32.18& 81.11 & 3.7 & 245.2 \\
& Mix-AdaLog (\!\!\cite{AdaLog_wu2025}+MP)$^*$ &66.23& 81.42 & 11.2 & 204.9 \\
& Mix-AQViT (ours)$^*$ &69.66& 82.16 & 11.3 & 202.1 \\
\hline 
\end{tabular}%
}
\end{table}

\subsection{Efficiency Analysis}

We evaluate the effectiveness of various methods in terms of data requirements, calibration time, and model inference throughput, as detailed in Table~\ref{tab:Efficiency_analysis_Data_Time_Throughput}, using a single NVIDIA 3090 GPU. All models presented in this analysis use 32 images for calibration. The table compares calibration times for 4/4-bit fixed and mixed-precision configurations. For 3/3-bit fixed and mixed-precision, the calibration times are nearly identical to those of the 4/4-bit configurations, with only marginal differences. Our fixed-precision frameworks, RQViT and AQViT, achieve calibration times and throughput comparable to RepQ-ViT and AdaLog, respectively, while significantly outperforming them in accuracy. For instance, AQViT achieves 74.99\% (4-bit) and 44.56\% (3-bit) on DeiT-S, outperforming AdaLog's 70.06\% (4-bit) and 22.82\% (3-bit), with similar calibration time and throughput. Additionally, our mixed-precision methods, Mix-RQViT and Mix-AQViT, retain the same model size and bit operations as their fixed-bit quantized counterparts while providing comparable throughput and substantial accuracy improvements. Similar trends are observed with the Swin-S model. 

\section{Conclusions}
\label{sec:conclusion}

In this paper, we propose Mix-QViT, a novel approach for post-training quantization and quantization-aware training with mixed-bit quantization for vision transformers. Mix-QViT introduces a comprehensive methodology for determining optimal bit allocations based on layer importance and quantization sensitivity scores, under constraints such as model size and bit operations. For PTQ, we introduce clipped channel-wise quantization for post-LayerNorm activations, which mitigates outliers, reduces inter-channel variation, and enhances model stability and performance. Extensive experiments demonstrate that Mix-QViT outperforms existing methods in low-bit PTQ and QAT, achieving significant accuracy gains with minimal impact on computational efficiency, making it highly effective for resource-constrained deployments.

\section*{Acknowledgments}
The authors acknowledge Research Computing at the Rochester Institute of Technology for providing computational resources and support that have contributed to the research results reported in this publication.




\end{document}